\documentclass[acmtog, authorversion, nonacm]{acmart}

\usepackage{booktabs} 

\usepackage[ruled,vlined,linesnumbered]{algorithm2e}

\SetAlFnt{\small}
\SetAlCapFnt{\small}
\SetAlCapNameFnt{\small}
\SetAlCapHSkip{0pt}

\acmJournal{TOG}

\setcopyright{none}

\makeatletter
\renewcommand\footnotetextauthorsaddresses[1]{}
\renewcommand\@copyrightpermission{}
\renewcommand\@formatdoi[1]{}
\makeatother

\begin{document}
\title{Context as Memory: Scene-Consistent Interactive Long Video Generation with Memory Retrieval}

\author{Jiwen Yu}
\authornote{Work done during an
internship at Kling Team, Kuaishou Technology}
\email{yujiwen.hk@connect.hku.hk}
\affiliation{
  \institution{The University of Hong Kong}
  \country{China}
}

\author{Jianhong Bai}
\authornotemark[1]
\email{cpurgicn@gmail.com}
\affiliation{
  \institution{Zhejiang University}
  \country{China}
}

\author{Yiran Qin}
\email{yiranqin@link.cuhk.edu.cn}
\affiliation{
  \institution{The University of Hong Kong}
  \country{China}
}

\author{Quande Liu}
\authornote{Corresponding authors}
\email{qdliu0226@gmail.com}
\author{Xintao Wang}
\email{xintao.wang@outlook.com}
\author{Pengfei Wan}
\email{wanpengfei@kuaishou.com}
\author{Di Zhang}
\email{zhangdi08@kuaishou.com}
\affiliation{
  \institution{Kling Team, Kuaishou Technology}
  \country{China}
}

\author{Xihui Liu}
\authornotemark[2]
\email{xihuiliu@eee.hku.hk}
\affiliation{
  \institution{The University of Hong Kong}
  \country{China}
}

\settopmatter{printacmref=false}

\begin{abstract}
Recent advances in interactive video generation have shown promising results, yet existing approaches struggle with scene-consistent memory capabilities in long video generation due to limited use of historical context. In this work, we propose \textbf{Context-as-Memory}, which utilizes historical context as memory for video generation. It includes two simple yet effective designs: (1) storing context in frame format without additional post-processing; (2) conditioning by concatenating context and frames to be predicted along the frame dimension at the input, requiring no external control modules. Furthermore, considering the enormous computational overhead of incorporating all historical context, we propose the \textbf{Memory Retrieval} module to select truly relevant context frames by determining FOV (Field of View) overlap between camera poses, which significantly reduces the number of candidate frames without substantial information loss. Experiments demonstrate that \textbf{Context-as-Memory} achieves superior memory capabilities in interactive long video generation compared to SOTAs, even generalizing effectively to open-domain scenarios not seen during training.
The link of our project page is \textcolor{red}{\url{https://context-as-memory.github.io/}}.
\end{abstract}

\begin{teaserfigure}
\centering
  \includegraphics[width=1\textwidth]{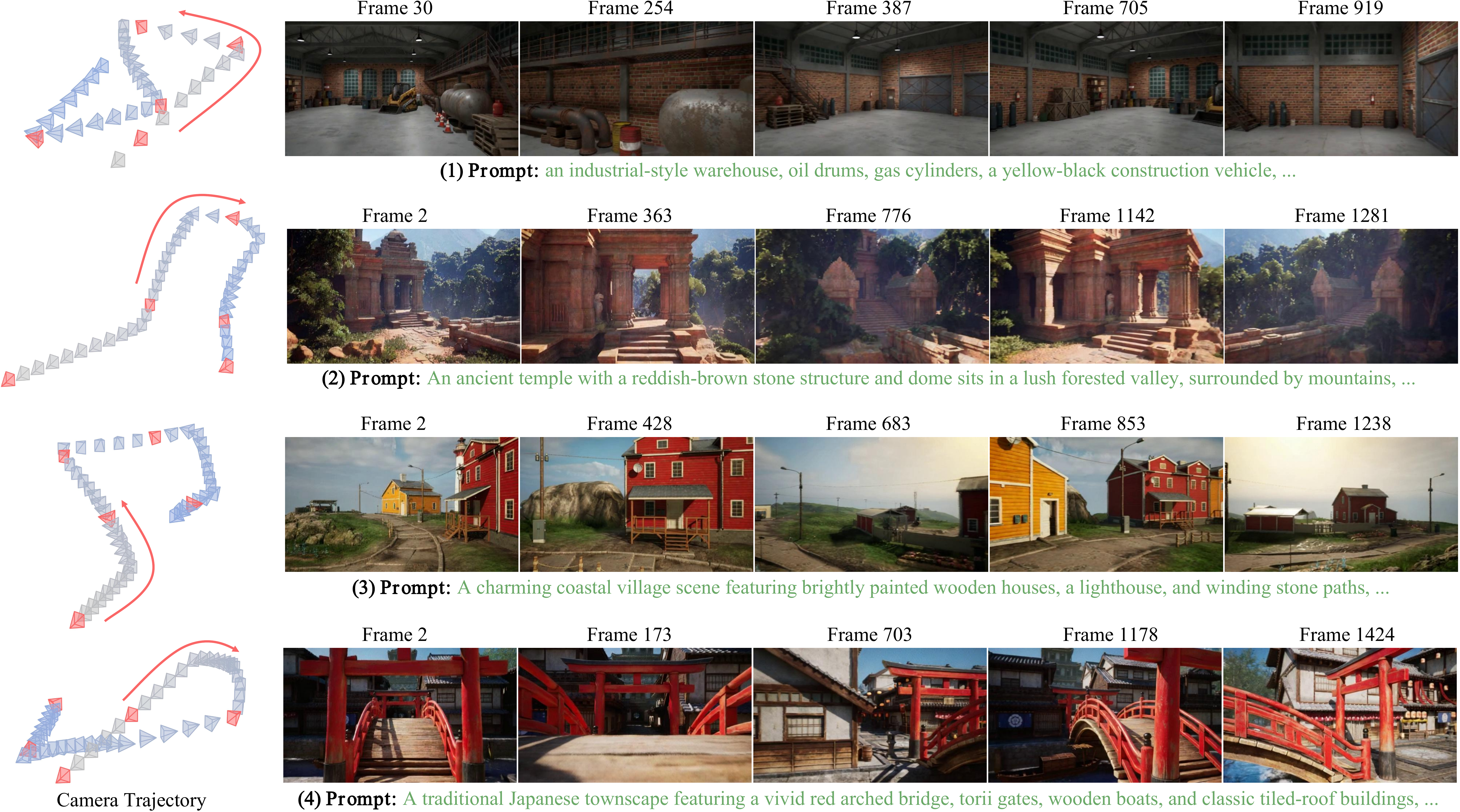}
  \caption{\textbf{Teaser Demonstration}. We propose \textbf{Context-as-Memory}, which utilizes history context as memory to guide the generation of new frames, thereby achieving scene-consistent long video generation. This figure shows key frames in generated videos under long camera trajectories. The cameras of the key frames are marked in \textcolor{red}{red}. It can be observed that different frames maintain good consistency when viewing the same scene from different viewpoints.}
  \label{fig:teaser}
\end{teaserfigure}

\maketitle
\newcommand{\todo}[1]{\textcolor{red}{TODO #1}}
\newcommand{\tocite}{\textcolor{red}{TO Cite}}
\newcommand{\modelname}{LVDM\xspace}

\section{Introduction}
\label{sec:intro}
Recent breakthroughs in video generation models~\cite{sora,runway,cogvideox,wan,hunyuanvideo} have shown remarkable progress. Due to their powerful generative capabilities developed through training on large-scale real-world datasets, these models are considered to have the potential to become world models capable of modeling reality~\cite{sora,video-new-language,unisim,worldsimbench}. 
Among various research directions in this field, interactive long video generation has emerged as a crucial one since many applications, such as gaming~\cite{genie2,gamengine,gamefactory} and simulation~\cite{gaia-1,vista,gaia-2}, require interactive long video generation, where the videos are generated in a streaming manner controlled by user interactions.
Recent works on long video generation~\cite{cdf,dfot,videopoet,emu3,nova,far,framepack} have significantly facilitated research in this field.

Despite these advances, current approaches still face significant challenges in terms of memory capabilities~\cite{igv-position,igv-survey}, which refers to a model's ability to maintain content consistency during continuous video generation, such as preserving the scene when the camera returns to a previously viewed location. Take Oasis~\cite{oasis} as an example: while it can generate lengthy Minecraft gameplay videos, even simple operations like turning left and then immediately right result in completely different scenes. This issue is prevalent across various state-of-the-art methods~\cite{gamengine,wham,dfot,gamefactory}, suggesting that while current approaches can generate videos of extended duration, they struggle to maintain coherent long-term memory of scene content and spatial relationships.

However, in our view, these methods' limitations in memory capabilities are not surprising. This is because when generating each new video frame, these methods can only predict based on a limited number of previous frames. For instance, Diffusion Forcing~\cite{cdf, dfot} can only utilize context from a fixed window of several dozen frames. While this setup works for video continuation, it fails to maintain long-term consistency. 
In the case of video generation, if each frame to be generated could reference all previously generated frames, the generative model could actively select and replicate relevant content from historical frames into the current frame being generated, thus it would be possible to maintain scene consistency in long videos. In other words, \textbf{all previously generated context frames serve as the memory.}

However, the idea of "all historical context as memory" seems intuitive but is impractical for three main reasons:
(1) Including all historical frames in computation would be extremely resource-intensive. 
(2) Processing all historical frames is computationally wasteful since only a small fraction is relevant to the current frame generation.
(3) Processing irrelevant historical frames adds noise that may hinder rather than help current frame generation.
Therefore, a reasonable approach is to retrieve a small number of relevant frames from historical context as conditions for current generation, which we call "\textbf{Memory Retrieval}".

In this work, we propose \textbf{Context-as-Memory} as a solution for scene-consistent interactive long video generation, which includes two simple yet effective designs: (1) Storage format: directly store generated context frames as memory, requiring no post-processing such as feature embedding extraction or 3D reconstruction; (2) Conditioning method: directly incorporate as part of the input through concatenation for context learning, without requiring additional control modules like external adapters or cross attention.
To effectively reduce unnecessary computational overhead and only condition on truly relevant context, we propose \textbf{Memory Retrieval}. Specifically, we introduce a rule-based approach based on camera trajectories. With a camera-controlled video generation model, we can annotate all context frames with camera information based on user's camera control. We can determine co-visibility by checking the FOV (Field of View) overlap based on camera poses at each timestamp along the trajectory, and then use this co-visibility relationship to decide which relevant frames to retrieve.
To implement this solution, we collected a new scene-consistent memory learning dataset using Unreal Engine 5, featuring long videos with precise camera annotations across diverse scenes and camera trajectories. The same regions are captured across different viewpoints and times, enabling both FOV-based retrieval and long-term consistency supervision.


Our main contributions can be summarized as follows:
\begin{itemize}
    \item We propose \textbf{Context-as-Memory}, highlighting the direct storage of frames as memory and conditioning via historical context learning for scene-consistent video generation.
    \item To effectively utilize relevant history frames while minimizing costs, we design \textbf{Memory Retrieval}, a specialized rule-based approach using FOV overlap of camera trajectory.
    \item We introduce a long, scene-consistent video dataset with precise camera annotations for memory training, featuring diverse scenes and captions.
    \item Our experiments show superior long video generation memory, significantly outperforming SOTAs and achieving effective memory even in unseen, open-domain scenarios.
\end{itemize}

\section{Related Work}
\subsection{Interactive Long Video Generation}
In following parts, we will review related work from four aspects:

\paragraph{Video Generation Model.} Video generation models can generate video sequences $\mathbf{x} = \{ x^0, x^1, \dots, x^t\}$, where $x^i$ indicates the $i$-th frame. The current mainstream model architecture is based on diffusion models~\cite{ddpm, sde, score,flow,rectified}, which excel in generating high-quality content and have been widely adopted in video generation~\cite{sora,cogvideox,kling,veo2,runway,vidu,wan,hunyuanvideo}. Other alternative architectures include next-token prediction~\cite{videogpt,videopoet,emu3} and various hybrid approaches~\cite{cdf,nova,mar}.

\paragraph{Controllable Video Generation.} This task can be formulated as $p(\mathbf{x}|c)$, where $c$ represents different types of control signals. The most representative control signals include: camera motion control~\cite{motionctrl,cameractrl,3dtrajmaster,recammaster, syncammaster}, and agent action control in games or simulators~\cite{gamengine,genie2,gamefactory,matrix,oasis}. These control signals greatly enhance user interactive experience, enabling free exploration in the created virtual worlds.

\paragraph{Streaming Video Generation.} Streaming video generation can condition on previously generated frames to continuously generate new video frames, which can be expressed as $p(x^0,x^1,...,x^n)=\prod_{i=0}^np(x^i|x^0,x^1,...,x^{i-1})$, where $x^i$ indicates the $i$-th frame. Representative approaches include Diffusion-based methods~\cite{cdf,dfot,gamefactory,far,framepack} and GPT-like next token prediction methods~\cite{videopoet,emu3,wham}. Diffusion-based methods generally achieve higher visual quality and faster sampling speed, thus we focus on diffusion models for long video generation in this work. Although these SOTA methods generally fail to generate long videos with scene-consistent memory, instead only producing long videos with short-term continuity.

\paragraph{Memory Capability for Video Generation.} Many related works' demos~\cite{gamengine,oasis,dfot,wham} have shown that current long video generation methods generally lack memory capability: while maintaining frame-to-frame continuity, the scenes continuously change. One potential approach~\cite{wonderjourney,viewcrafter,see3d,gen3c} is to leverage 3D reconstruction to build explicit 3D representations from generated videos, then render initial frames from these 3D representations as conditions for new video generation. However, this method is limited by the accuracy and speed of 3D reconstruction, particularly in continuously expanding large scenes where accumulated 3D reconstruction errors become intolerable. Moreover, these works focus on 3D generation and merely borrow priors from video generation models, which differs from our scope. WorldMem~\cite{worldmem} attempts to implement memory by injecting historical frames through cross attention, and has been validated on video lengths of around $10$ seconds in Minecraft scenarios.

\subsection{Context Learning for Video Generation}
Recently, some works~\cite{lct,far,framepack} have begun to explore the role of long-context in video generation. LCT~\cite{lct} performs long-context tuning on pre-trained single-shot video diffusion models to achieve consistency in multi-shot video generation. FAR~\cite{far} proposes Long-Term and Short-Term context windows to condition video generation models for long video generation. FramePack~\cite{framepack} introduces a hierarchical method to compress context frames into a fixed number of frames as conditioning for video generation models to achieve long video generation. However, their compression method loses too much information from temporally distant frames. In this work, we further highlight the significance of context, emphasizing that all history context serves as memory for scene-consistent long video generation.

\section{Method}
\label{sec:method}
As discussed in Section \ref{sec:intro}, we propose that historical context frames can serve as memory for scene-consistent interactive long video generation. This section will detail how we implement this approach. Specifically:
Sec.~\ref{subsec:preliminary} introduces preliminaries.
Sec.~\ref{subsec:inject_context} describes how to inject context frames as conditions for video generation.
Sec.~\ref{subsec:acc} presents our Memory Retrieval method, which selects most relevant context frames to guide the generation of new frames. This section includes alternative approaches and our proposed search method based on camera trajectories.
Sec.~\ref{subsec:data} introduces our long video dataset collected using Unreal Engine 5, which features precise camera pose annotations, diverse scenes, and caption annotations.

\subsection{Preliminaries}
\label{subsec:preliminary}
\begin{figure}[t]
  \centering
  \includegraphics[width=1\linewidth]{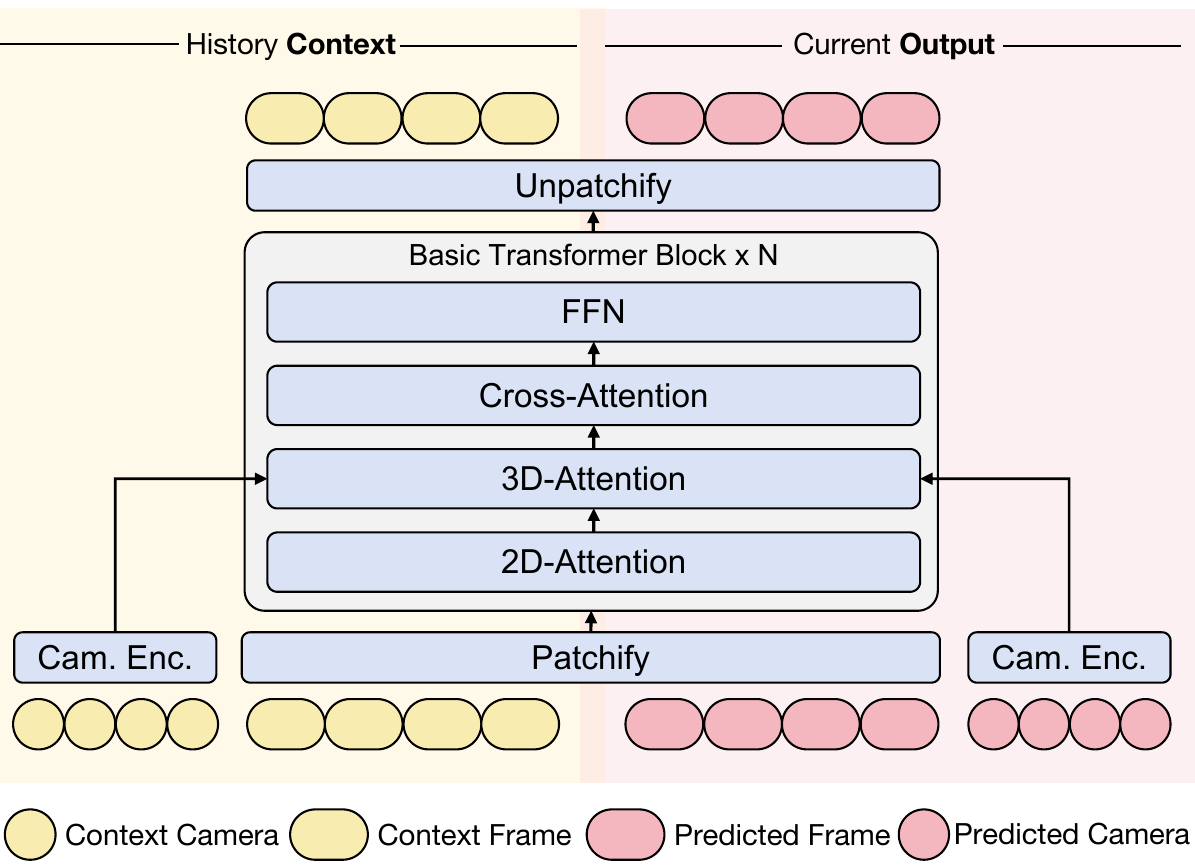}
  \vspace{-0.6cm}
  \caption{
  \textbf{Model Architecture.} We concatenate the context to be conditioned and the predicted frames along the frame dimension. This method of injecting context is simple and effective, requiring no additional modules.
  }
  \vspace{-0.2cm}
\label{fig:arch} 
\end{figure}
\begin{figure*}[ht]
  \centering
  \includegraphics[width=1\linewidth]{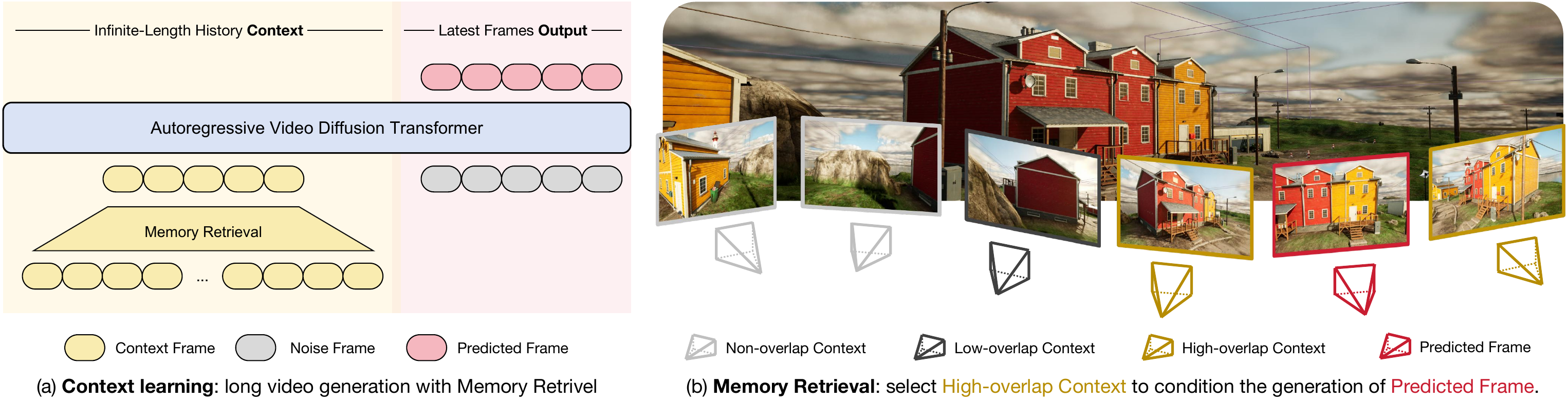}
  \vspace{-0.6cm}
  \caption{
  \textbf{Method Demonstration.} (a) We propose \textbf{Context-as-Memory}, where all historical context frames serve as memory conditions in the generation of predicted frames, with Memory Retrieval extracting relevant information from all context frames. (b) Our proposed Memory Retrieval method is a search algorithm based on camera trajectories. It selects relevant frames by evaluating the overlap between camera views of different frames.
  }
  \vspace{-0.2cm}
\label{fig:framework} 
\end{figure*}
\paragraph{Full-Sequence Text-to-Video Base Model.} Our work is based on a full-sequence text-to-video model, specifically, a latent video diffusion model consisting of a causal 3D VAE~\cite{vae} and a Diffusion Transformer (DiT)~\cite{dit}. Each DiT block sequentially consists of spatial (2D) attention, spatial-temporal (3D) attention, cross-attention, and FFN modules.
Let $\mathbf{x}$ represent a sequence of video frames, the Encoder of 3D VAE compresses it temporally and spatially to obtain the latent representation $\mathbf{z}=Encoder(\mathbf{x})$.
With a temporal compression factor of $r$, the original $1+nr$ frames of $\mathbf{x}=\{x^0,x^1,...x^{nr}\}$ are compressed into $1+n$ latents of $\mathbf{z}=\{z^0,z^1,...,z^{n}\}$.
During training, random Gaussian noise $\boldsymbol{\epsilon}\sim\mathcal{N}(\mathbf{0},\mathbf{I})$ is added to the clean latent $\mathbf{z}_0$ to obtain noisy latent $\mathbf{z}_t$ at timestep $t$. The network $\epsilon_{\phi}(\cdot)$ is trained to predict the added noise, with the following loss function:
\begin{equation}
    \mathcal{L}(\phi)=\mathbb{E}[||\boldsymbol{\epsilon}_{\phi}(\mathbf{z}_t,\mathbf{p},t)-\boldsymbol{\epsilon}||],
\end{equation}
where $\phi$ represents the parameters and $\mathbf{p}$ is the given text prompt.
Then we can use the predicted noise $\boldsymbol{\epsilon}_{\phi}$ to denoise the noisy latent.
During inference, a clean latent $\mathbf{z}$ can be sampled from a randomly sampled Gaussian noise, then the Decoder of 3D VAE decodes it into video sequence $\mathbf{x}=Decoder(\mathbf{z})$.

\paragraph{Camera-Conditioned Video Generation.} In our work, we incorporate camera control mechanisms~\cite{motionctrl,recammaster} into the video generation model to implement interactive video generation. 
By providing camera trajectories as conditioning for video generation, we can know the camera poses of each context frame in advance. 
Let $\mathbf{cam}$ represent the camera poses, where $f$ denotes the total number of frames. Following the mechanism proposed in ReCamMaster~\cite{recammaster}, in order to inject $\mathbf{cam}=[R,t]\in\mathbb{R}^{f\times(3\times4)}$, we first map it to the same dimension as the model's feature channels through a camera encoder $\mathcal{E}_{c}(\cdot)$, followed by adding them together:
\begin{equation}
    \mathbf{F}_i = \mathbf{F}_o + \mathcal{E}_{c}(\mathbf{cam}),
\end{equation}
where $\mathbf{F}_o$ is the output of spatial attention module, $\mathbf{F}_i$ is the input of 3D attention module and $\mathcal{E}_c(\cdot)$ is one layer of MLP with $\phi_{MLP}$ as learnable parameters. During the training of camera control, we use the original diffusion loss as follows:
\begin{equation}
    \mathcal{L}_{\mathbf{cam}}(\phi,\phi_{MLP})=\mathbb{E}[||\boldsymbol{\epsilon}_{\phi,\phi_{MLP}}(\mathbf{z}_t,\mathbf{p},\mathbf{cam},t)-\boldsymbol{\epsilon}||].
\end{equation}

\subsection{Context Frames Learning Mechanism for Memory}
\label{subsec:inject_context}
Suppose the latent of context that needs to be conditioned is $\mathbf{z}^{c}$, and we need to learn the conditional denoiser $p(\mathbf{z}_{t-1}|\mathbf{z}_t,\mathbf{z}^{c})$.
Considering that the context grows continuously during the generation process (i.e., the context is variable-length), methods designed for single-frame or fixed-length frame conditions, such as Adapter~\cite{controlnet,t2iadapter} and channel-wise concatenation~\cite{dynamicrafter}, are not applicable. Similar to ReCamMaster~\cite{recammaster}, we propose to inject context through concatenation along the frame dimension (shown in Fig.~\ref{fig:arch}), which can flexibly support variable-length context conditions. Specifically, the clean context latents $\mathbf{z}^c$ participate equally with the noisy predicted latents $\mathbf{z}_t$ in the attention computation within DiT Blocks. During output, we only update the noisy latents $\mathbf{z}_t$ using the predicted noise $\boldsymbol{\epsilon}_{\phi}(\{\mathbf{z}_t,\mathbf{z}^c\},\mathbf{p},t)$ while keeping the clean context latents $\mathbf{z}^c$ unchanged.

Another challenge is how to handle positional encoding along the frame dimension in video diffusion models after context frame expansion. Since our method is based on a pre-trained full-sequence text-to-video model, to preserve the original model's generation capability and facilitate easier adaptation to the context-conditioned generation setting, we maintain the same positional encoding for predicted latents $\mathbf{z}_t$ as in the pre-training phase, while assigning new positional encodings to the newly conditioned context latents $\mathbf{z}^c$. Our base model employs RoPE~\cite{rope}, which can conveniently adapt to variable-length position encodings.

\subsection{Memory Retrieval}
\label{subsec:acc}
As analyzed in Sec.~\ref{sec:intro}, including all context frames in computation is impractical due to computational overhead and may introduce irrelevant information that causes interference. A reasonable approach is to filter out valuable frames from the context, specifically frames that share overlapping visible regions with the frames to be generated. To this end, we propose \textbf{Memory Retrieval} to accomplish this task as shown in Fig.~\ref{fig:framework} (a). Below, we first introduce several alternative implementation methods, followed by our solution.

\paragraph{Alternative method \#1: random selection.} 
A baseline randomly selects frames from context. This works well in early generation when context size is small, as adjacent frames' natural redundancy reduces the risk of missing important information. However, with hundreds of context frames, random selection fails to identify valuable frames.

\paragraph{Alternative method \#2: neighbor frames within a window.} 
Another approach selects consecutive recent frames within a window near the current predicted frames. While common in existing methods~\cite{dfot,oasis,gamefactory}, this has key limitations. First, adjacent frames' redundancy means multiple consecutive frames add little new information beyond the most recent frame. Second, ignoring temporally distant frames prevents awareness of previously seen scenes, leading to continuous generation of new scenes and ultimately breaking scene consistency.

\paragraph{Alternative method \#3: hierarchical compression.} FramePack~\cite{framepack} proposes a hierarchical compression method for context frames into a minimal set (e.g., 2-3 frames). For two-frame compression, it allocates space proportionally: the most recent frame gets one full frame, the second most recent gets half, the third gets a quarter, and so on, totaling two frames. While achieving high compression, this exponential decay significantly loses historical information. Though the authors suggest manually preserving certain key frames uncompressed, they don't specify the selection criteria.

\paragraph{Our method: camera-trajectory-based search.}
The fundamental limitation of these methods lies in their inability to identify truly valuable frames from the large number of context frames. They either introduce many redundant frames or lose too much useful information, especially from the old frames that are temporally distant.
We leverage the known camera trajectory of the context to search for valuable frames, specifically those that share high-overlap visible regions with predicted frame as shown in Fig.~\ref{fig:framework} (b).

The first question is how to obtain the camera trajectory of the context video. Since we have introduced camera control into our video generation model in Sec.~\ref{subsec:preliminary}, these context frames are generated with user-provided camera poses. These conditioning camera poses can serve as camera annotations for the generated context, eliminating the need for an additional camera pose estimator.

\begin{figure}[t]
  \centering
  \includegraphics[width=1\linewidth]{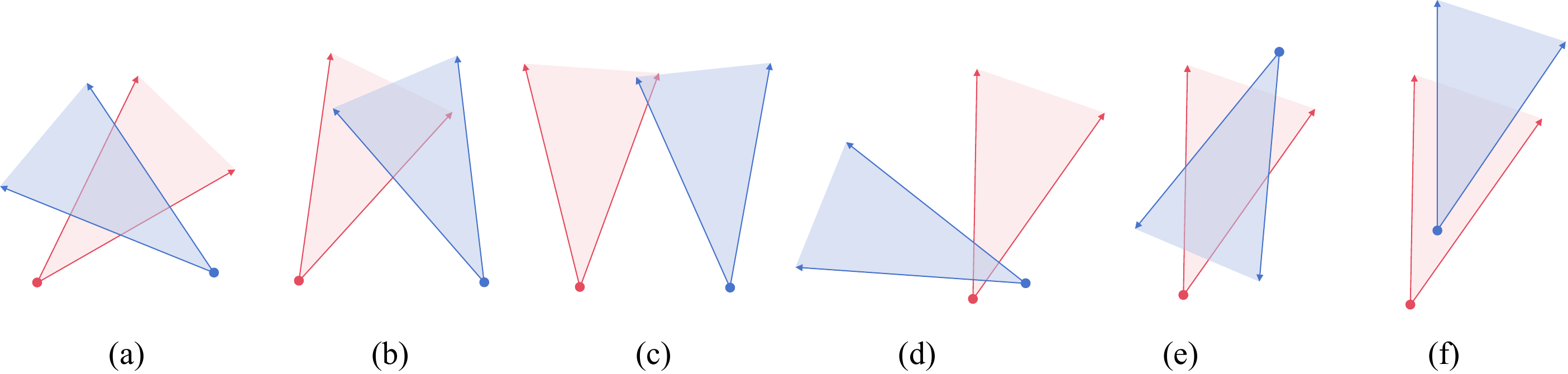}
  \vspace{-0.6cm}
  \caption{
  \textbf{Examples of FOV Overlap.} 
  We simplify FOV overlap detection to checking intersections between four rays from two camera origins. A practical rule that works for most cases requires: both left and right ray pairs intersect (a, b). However, we must filter out cases where intersection points are either too near (d) or too distant (c) from cameras. While this rule may not cover all scenarios and some corner cases exist (e, f), occasional missed or incorrect candidates don't substantially affect overall performance.
  }
  \vspace{-0.2cm}
\label{fig:fov} 
\end{figure}
The second question is how to determine co-visibility between two frames given their camera poses. We attempt to determine this by checking if there is an overlapping region between the fan-shaped areas corresponding to the Fields of View (FOV) of the two cameras. Specifically, since we restrict camera movement to the XY plane, we only need to consider the left and right rays shooting from each camera's origin. By checking the intersection of these four rays from two cameras, we can quickly determine the FOV overlap as shown in Fig.~\ref{fig:fov}. Additionally, we calculate the distance between the predicted frame's camera and the calculated intersection points to eliminate cases where the cameras are too far apart (which typically indicates no actual overlap or very small overlap). This FOV overlap detection is not perfect, as it may fail in cases with occlusions. However, this method effectively reduces the number of candidate context frames.

The final question is: after FOV co-visibility filtering, if the number of filtered frames still exceeds the context condition limit, how should we further filter them? A baseline approach would be random selection, but we also provide some more insightful strategies:
(1) Considering the redundancy between adjacent frames, we randomly select only one frame from each group of consecutive frames in the filtered context. This design is highly effective, significantly reducing the number of candidate frames while preserving most of the valuable information.
(2) Building upon the first strategy, we can additionally select a few context frames that are furthest apart either spatially or temporally. This helps to supplement potentially missing long-term information (both spatial and temporal). However, in most cases, this additional selection may not be necessary.

\paragraph{Implementation details in training and inference.} Assume the maximum number of retrieved context frames is $k$. 
During training, we read a long ground truth video (containing thousands of frames) and randomly select a segment as the sequence to be predicted. We then apply our Memory Retrieval method to select $k-1$ context frames from the remaining frames. The overlapping relationships between frames have been pre-computed, eliminating the need for repeated calculations. The first frame of the prediction sequence is also included as an additional context frame to ensure video continuity. Additionally, there is a $10\%$ probability during training that only the recent context frame is used, simulating the beginning of long video generation where no context frames are available. 
During inference, for each video segment to be predicted, we search $k-1$ context frames from the previously generated frames using FOV-based Memory Retrieval and add the most recently generated frame to the context.
The training and inference procedures are outlined in Algorithm \ref{alg:training} and \ref{alg:inference}, respectively.


\begin{algorithm}[t]
\caption{Training Process of \textbf{Context-as-Memory}}
\label{alg:training}
\KwIn{Video sequence $\mathcal{X}$ and camera annotations $\mathcal{C}$ in training dataset, context size $k$}

\While{not converged}{
    Randomly select predicted video sequence $\mathbf{x}_0$ from $\mathcal{X}$\;
    Retrieve $k$ frames as context $\mathbf{x}^{c}$\;
    Obtain camera poses $\{\mathbf{cam}_0, \mathbf{cam}^c\}$ for $\{\mathbf{x}_0, \mathbf{x}^c\}$ from $\mathcal{C}$\;
    Obtain latent embeddings $\{\mathbf{z}_0, \mathbf{z}^c\} \gets \text{Encoder}(\{\mathbf{x}_0, \mathbf{x}^c\})$\;
    Sample $t \sim U(1, T)$ and $\epsilon \sim \mathcal{N}(0, \mathbf{I})$, then corrupt $\mathbf{z}_0$ to $\mathbf{z}_t$\;
    Train $p(\mathbf{z}_{t-1} \mid \mathbf{z}_t, \mathbf{z}^c, \mathbf{cam}_0, \mathbf{cam}^c, t)$ using diffusion loss\;
}
\end{algorithm}

\begin{algorithm}[t]
\caption{Inference Process of \textbf{Context-as-Memory}} \label{alg:inference}
\KwIn{Initial frame set $\mathcal{X} = \{\mathbf{x}_\text{init}\}$ and camera poses $\mathcal{C} = \{\mathbf{cam}_\text{init}\}$}
\KwOut{Generated video sequence $\mathcal{X}$}

\While{generation not finished}{
    User provides next target camera pose $\mathbf{cam}^t$\;
    Retrieve context frames $\mathbf{x}^c \subset \mathcal{X}$ and $\mathbf{cam}^c \subset \mathcal{C}$ by checking FOV overlap with $\mathbf{cam}^t$\;
    Compute context latent $\mathbf{z}^c \gets \text{Encoder}(\mathbf{x}^c)$\;
    Sample noise $\epsilon \sim \mathcal{N}(0, \mathbf{I})$ and infer latent $\mathbf{z}^t \sim p(\mathbf{z}^t \mid \epsilon, \mathbf{z}^c, \mathbf{cam}^t, \mathbf{cam}^c)$\;
    Decode generated frames $\mathbf{x}^t \gets \text{Decoder}(\mathbf{z}^t)$\;
    Append $\mathbf{x}^t$ to $\mathcal{X}$ and $\mathbf{cam}^t$ to $\mathcal{C}$\;
}
\end{algorithm}

\subsection{Data Collection}
\label{subsec:data}
To validate our method, we require long video datasets with camera pose annotations. However, currently available datasets with camera pose information typically consist of short video clips~\cite{realestate, recammaster}. To obtain long-duration data with precise camera annotations, we utilized a simulation environment, specifically Unreal Engine 5. We generated randomized camera trajectories navigating through different scenes and rendered corresponding long videos. Our dataset comprises 100 videos of 7,601 frames each, featuring 12 distinct scene styles, with captions annotated by a multimodal LLM~\cite{minicpm} every 77 frames. To simplify the problem while still effectively validating our method, we constrained the camera trajectory's position changes to a 2D plane and limited rotation to only around the z-axis, which still presents sufficient complexity for camera trajectory control.
Additional details about the dataset are provided in the supplementary materials.

\section{Experiments}
\begin{figure*}[t]
  \centering
  \includegraphics[width=1\linewidth]{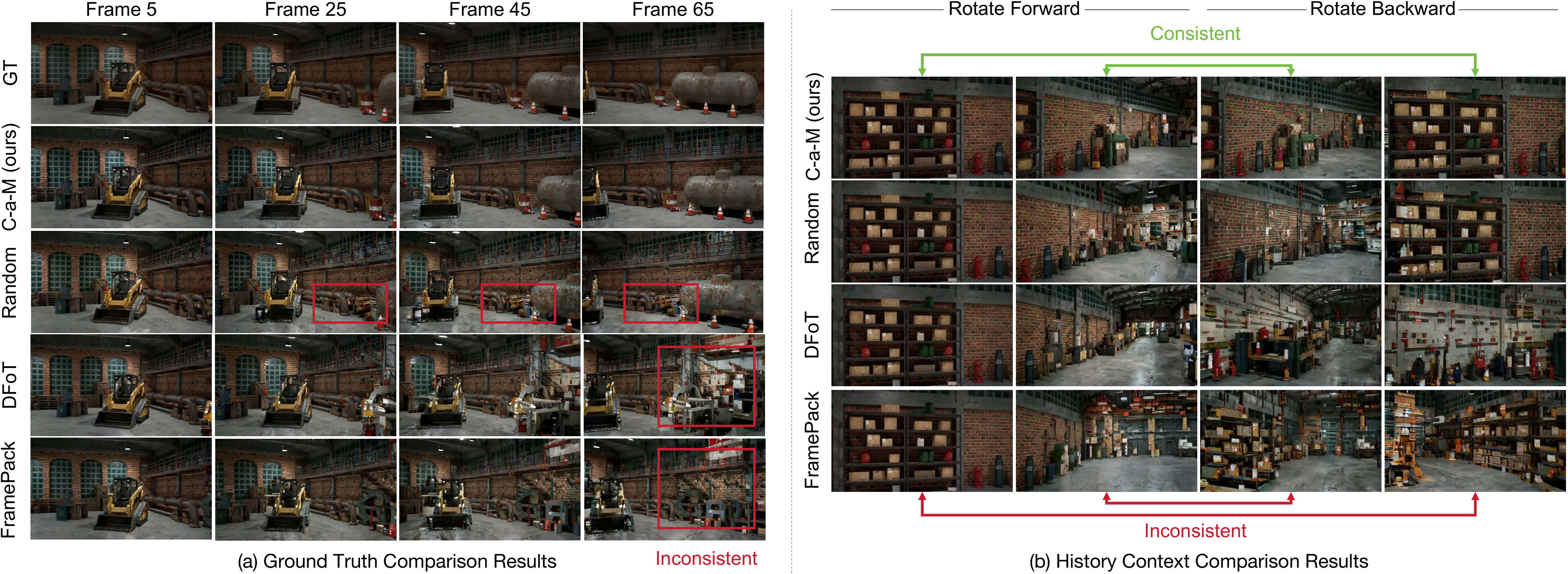}
  \vspace{-0.6cm}
  \caption{
  \textbf{Qualitative Comparison Results.} Among them, Context-as-Memory demonstrated the best memory capabilities and the highest visual quality, indicating the effectiveness of sufficient context information conditioning.
  Other methods exhibit scene inconsistency issues due to limited context utilization.
  }
  \vspace{-0.2cm}
\label{fig:comparison} 
\end{figure*}
\begin{table*}[t]

\tabcolsep=0.09cm
\center
\begin{tabular}{c | c c c c | c c c c }
\toprule
& \multicolumn{4}{c}{Ground Truth Comparison} & \multicolumn{4}{c}{History Context Comparison} \\
  Methods  & PSNR$\uparrow$ & LPIPS$\downarrow$ & FID$\downarrow$ & FVD$\downarrow$ & PSNR$\uparrow$ & LPIPS$\downarrow$ & FID$\downarrow$ & FVD$\downarrow$ \\
  \hline
  \hline
  1st Frame as Context & 15.72 & 0.5282 &  127.55 & 937.51 & 14.53 & 0.5456 &  157.44 & 1029.71 \\
  1st Frame + Random Context & 17.70 & 0.4847 &  115.94 & 853.13 & 17.07 & 0.3985 &  119.31 & 882.36 \\
  DFoT~\cite{dfot} & 17.63 & 0.4528 &  112.96 & 897.87 & 15.70 & 0.5102 &  121.18 & 919.75 \\
  FramePack~\cite{framepack} & 17.20 & 0.4757 &  121.87 & 901.58 & 15.65 & 0.4947 &  131.59 & 974.52 \\
  \textbf{Context-as-Memory (Ours)} & \textbf{20.22} & \textbf{0.3003} &  \textbf{107.18} & \textbf{821.37} & \textbf{18.11} & \textbf{0.3414} &  \textbf{113.22} & \textbf{859.42} \\
  
\bottomrule

\end{tabular}
\caption{
\textbf{Quantitative Comparison results.} Due to learning abundant context, Context-as-Memory demonstrates the best memory capabilities and highest quality of generated videos. In contrast, DFoT~\cite{dfot} and FramePack~\cite{framepack}, which can only utilize the most recent contexts, show relatively inferior performance, even worse than random context selection. This is because although random selection cannot guarantee the selection of useful information, on average it tends to obtain more information compared to methods that only learn from the most recent context.
}
\vspace{-0.6cm}
\label{table:comp}
\end{table*}
\subsection{Experiment Settings}
\paragraph{Implementation Details.}
Our method is implemented on an internal 1B-parameter pre-trained text-to-video Diffusion Transformer, developed for research purposes. The resolution of generated videos is $640\times 352$. The model supports generation of $77$-frame videos, with a temporal compression ratio of $4$ in the causal 3D VAE, resulting in $20$-frame video latents generation. We set the context size to $20$, meaning $20$ RGB frames are selected as context. Since these frames lack temporal continuity, they are individually compressed using the causal 3D VAE, also resulting in $20$ frames of video latents. The model was trained on our collected dataset for over $10,000$ iterations with a batch size of $64$ on $8$ NVIDIA A100 GPUs. During sampling, we employ Classifier-Free Guidance~\cite{cfg} for text prompts, with $50$ sampling steps.

\paragraph{Evaluation Methods.}
To evaluate our method, we held out $5\%$ of the dataset containing diverse scenes for testing. Our evaluation metrics include: 
(1) \textbf{FID and FVD} for video quality assessment; 
(2) \textbf{PSNR and LPIPS} for quantifying memory capability through pixel-wise differences between frames. 
Given the lack of memory evaluation methods, we propose two approaches: (1) \textbf{Ground truth comparison}: evaluating whether predicted frames match ground truth based on context selected from ground truth frames; (2) \textbf{History context comparison}: comparing newly generated frames with previously generated ones in long video sequences. This second approach provides stronger evidence of memory capability as it evaluates consistency in newly generated content. In our implementation, we test on simple trajectories where the camera rotates n degrees and returns, allowing easy identification of corresponding frames for PSNR/LPIPS calculation.

\subsection{Comparison Results}
In this section, we evaluate video generation memory capabilities across baseline methods, SOTA approaches, and our Context-as-Memory. The compared methods include:
(1) Single-frame context using the first frame;
(2) Multi-frame context using the first frame plus random historical frames;
(3) Diffusion Forcing Transformer (DFoT)~\cite{dfot}, using a fixed-size window of most recent frames;
(4) FramePack~\cite{framepack}, which hierarchically compresses previous context into two frames, with each frame's height or width halved compared to its predecessor. While theoretically supporting all historical frames, compression becomes impractical after several frames as latent size reduces to $1\times 1$.
For fair comparison, all methods were implemented on our base model and dataset with identical training configurations and iterations.
Results are presented in Table~\ref{table:comp} and Figure~\ref{fig:comparison}.


PSNR and LPIPS metrics demonstrate our Memory-as-Context's advantages over other approaches. It effectively retrieves and utilizes useful context information, while other methods have limited context access. Random context selection outperforms DFoT and FramePack, possibly because although it cannot guarantee selecting useful context, it still performs better on average than methods limited to only recent frames.
DFoT and FramePack's performance limitations stem from adjacent frame redundancy. Despite accessing dozens of recent frames, the inherent redundancy limits effective information utilization. FramePack's exponential information decay further weakens its memory capabilities compared to DFoT.

Moreover, FID and FVD show our Context-as-Memory achieves the best generation quality among all methods. Sufficient context conditioning not only enhances memory but also improves generation quality by reducing error accumulation in long videos. This improvement stems from two factors: (1) context provides stronger conditional guidance by reducing generation uncertainty, and (2) earlier generated frames used as context contain fewer accumulated errors, helping minimize error propagation in new frames.

Additionally, History Context Comparison proves more challenging than Ground Truth Comparison. Even with simple "rotate forward and rotate backward" trajectories, the performance gaps between methods are significant. DFoT and FramePack can only utilize the most recent context, causing them to continuously generate new content. Only by having access to global context and extracting useful relevant information from it can memory-aware new video generation be achieved.

\subsection{Ablation Study}
\begin{table}[t]

\tabcolsep=0.09cm
\center
\begin{tabular}{c | c c | c c | c}
\toprule
& \multicolumn{2}{c}{GT Comp.} & \multicolumn{2}{c}{HC Comp.} &\\
  Context Size  & PSNR$\uparrow$ & LPIPS$\downarrow$ & PSNR$\uparrow$ & LPIPS$\downarrow$  & Speed (fps)$\uparrow$ \\
  \hline
  \hline
   1 & 15.72 & 0.5282 & 14.53 & 0.5456 & 1.60 \\
   5 & 17.37 & 0.4825 & 15.97 & 0.5063 & 1.40 \\
   10 & 19.14 & 0.3554 & 17.75 & 0.3985 & 1.20 \\
   20 & 20.22 & 0.3003 & 18.11 & 0.3414 & 0.97 \\ 
   30 & 20.31 & 0.3137 & 18.19 & 0.3319 & 0.79 \\
\bottomrule

\end{tabular}
\caption{\textbf{Ablation of Context Size.} Larger context sizes contain more useful information and lead to better memory capability, but also incur higher computational overhead, necessitating an optimal trade-off choice.}
\vspace{-0.6cm}
\label{table:ablation_context}
\end{table}
\paragraph{Ablation of Context Size.} 
We studied how context size affects memory capability. Larger contexts theoretically provide more useful information, improving memory performance as shown in Tab.~\ref{table:ablation_context}. However, this comes with increased computational cost and slower generation speed. When context size reaches $30$, there's a notable speed drop compared to size $1$. Balancing performance and speed, a context size of $20$ offers a good trade-off. Future improvements in context compression techniques may help reduce the optimal context size further.




\begin{table}[t]

\tabcolsep=0.09cm
\center
\begin{tabular}{c | c c | c c }
\toprule
& \multicolumn{2}{c}{GT Comp.} & \multicolumn{2}{c}{HC Comp.}\\
  Strategy  & PSNR$\uparrow$ & LPIPS$\downarrow$ & PSNR$\uparrow$ & LPIPS$\downarrow$   \\
  \hline
  \hline
Random & 17.70 & 0.4847 & 17.07 & 0.3985 \\
  FOV+Random & 19.17 & 0.3825 & 17.47 & 0.3896 \\
  FOV+Non-adj & 20.11 & 0.3075 & \textbf{18.19} & 0.3571 \\
  FOV+Non-adj+Far-space-time & \textbf{20.22} & \textbf{0.3003} & 18.11 & \textbf{0.3414} \\

\bottomrule

\end{tabular}
\caption{\textbf{Ablation of Memory Retrieval Strategy.} The filtering methods of "FOV" and "Non-adj" (where only one frame from continuous frame sequences is selected as a candidate) effectively filter out useless and redundant information, leading to significant improvements in memory capability.}
\vspace{-0.6cm}
\label{table:ablation_retrieval}
\end{table}
\paragraph{Ablation of Memory Retrieval Strategy.} We ablated different memory retrieval strategies to analyze their effects. "Random" refers to randomly selecting context; "FOV+Random" means first filtering using the FOV-based method, then randomly selecting from the remaining candidates; "Non-adj" means only one frame from continuous frame sequences will be selected as a candidate; "far-space-time" means frames that are more distant in time or space are more likely to be selected. The results in Tab.~\ref{table:ablation_retrieval} demonstrate the effectiveness of "FOV" and "Non-adj" methods in removing useless and redundant information, which significantly increases the probability of selecting useful context and thereby enhances memory capability. The impact of "Far-space-time" is relatively minor.

\subsection{Open-Domain Results}
\begin{figure}[t]
  \centering
  \includegraphics[width=1\linewidth]{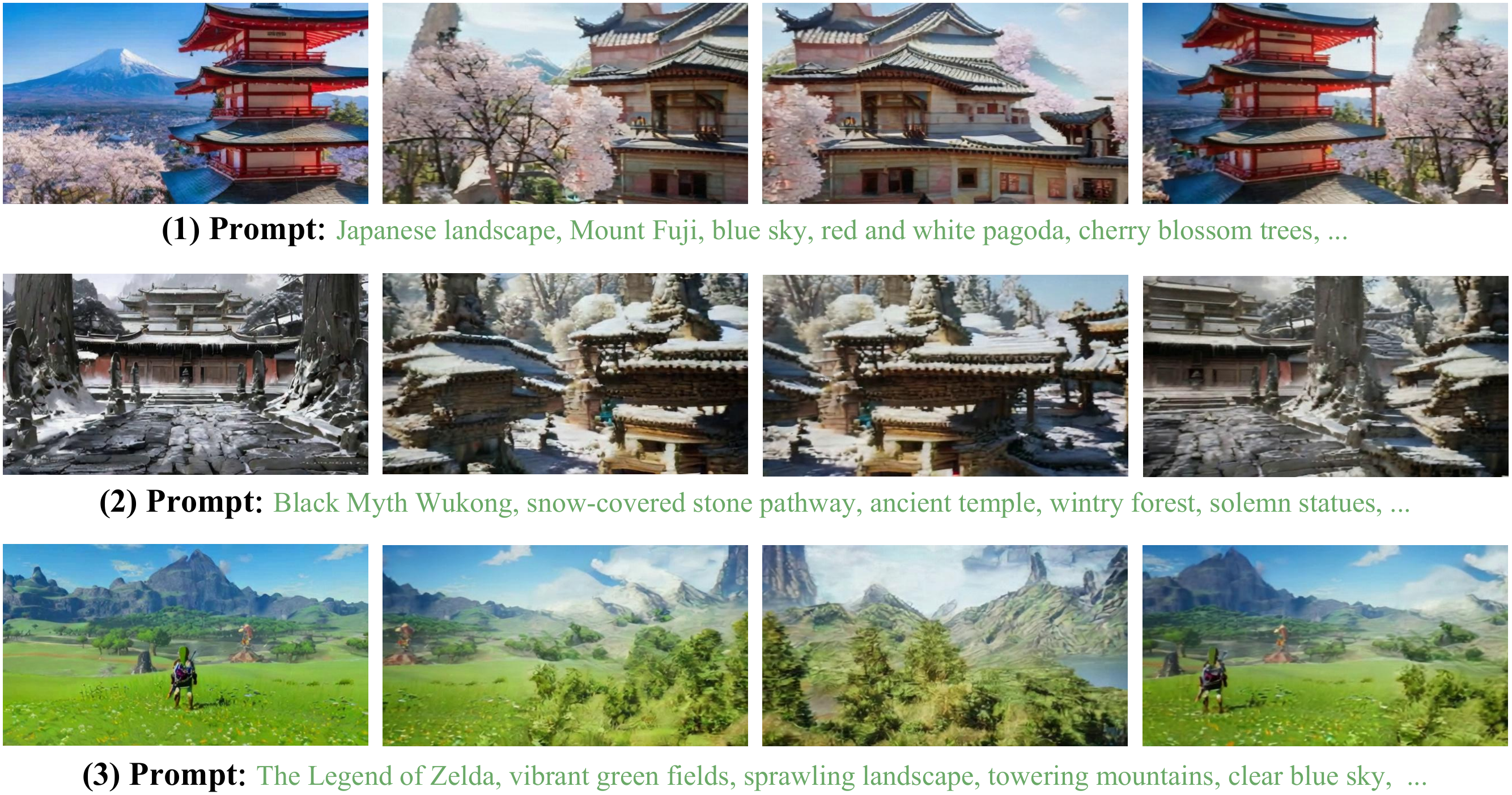}
  \vspace{-0.6cm}
  \caption{
  \textbf{Open-Domain Results.} We collected open-domain images from the internet and used them as the first frame to generate subsequent long videos. Under the trajectory of "rotate away and rotate back," even when generating new content, it still demonstrates good memory capability.
  }
  \vspace{-0.2cm}
\label{fig:open-domain} 
\end{figure}
Due to our diverse training dataset and the various visual priors learned by our base model during pre-training, our method has the potential to generalize to open-domain scenarios not present in the training set. We selected images of different styles from the internet and used them as the first frame to generate long videos. We validated using the trajectory of "rotate away and rotate back," which is suitable for verifying memory consistency in generated content. Results in Fig.~\ref{fig:open-domain} demonstrate that our method indeed possesses good memory capability in open-domain scenarios.


\section{Conclusion}
In this work, we propose \textbf{Context-as-Memory}, highlighting that using historical generated frames as memory is key to achieving scene-consistent long video generation. Our method design is simple yet effective, directly saving context frames as memory and inputting the context together with the predicted frame as conditions. Furthermore, to avoid high computational overhead caused by lengthy context, we propose \textbf{Memory Retrieval} to dynamically select truly valuable context based on the predicted video frames. 
\paragraph{Limitations and Future Work.}
Although our method has made significant progress in achieving memory capability for long video generation, several limitations remain: (1) Our method is limited to static scenes, while memory retrieval for dynamic scenes poses greater challenges; (2) In complex scenarios, particularly those with multiple occlusions (e.g., interconnected indoor rooms), FOV overlap may struggle to effectively identify truly relevant context frames; (3) The inherent error accumulation problem in long video generation persists, which currently can only be addressed through larger datasets, more extensive training, and more powerful base models. In the future, we will continue to develop memory capabilities for open-domain long video generation on larger-scale base models, supporting more complex trajectories, broader scene ranges, and longer generation sequences.

\bibliographystyle{ACM-Reference-Format}
\bibliography{sample-bibliography}

\appendix
\begin{figure*}[t]\centering
\includegraphics[width=0.6\textwidth]{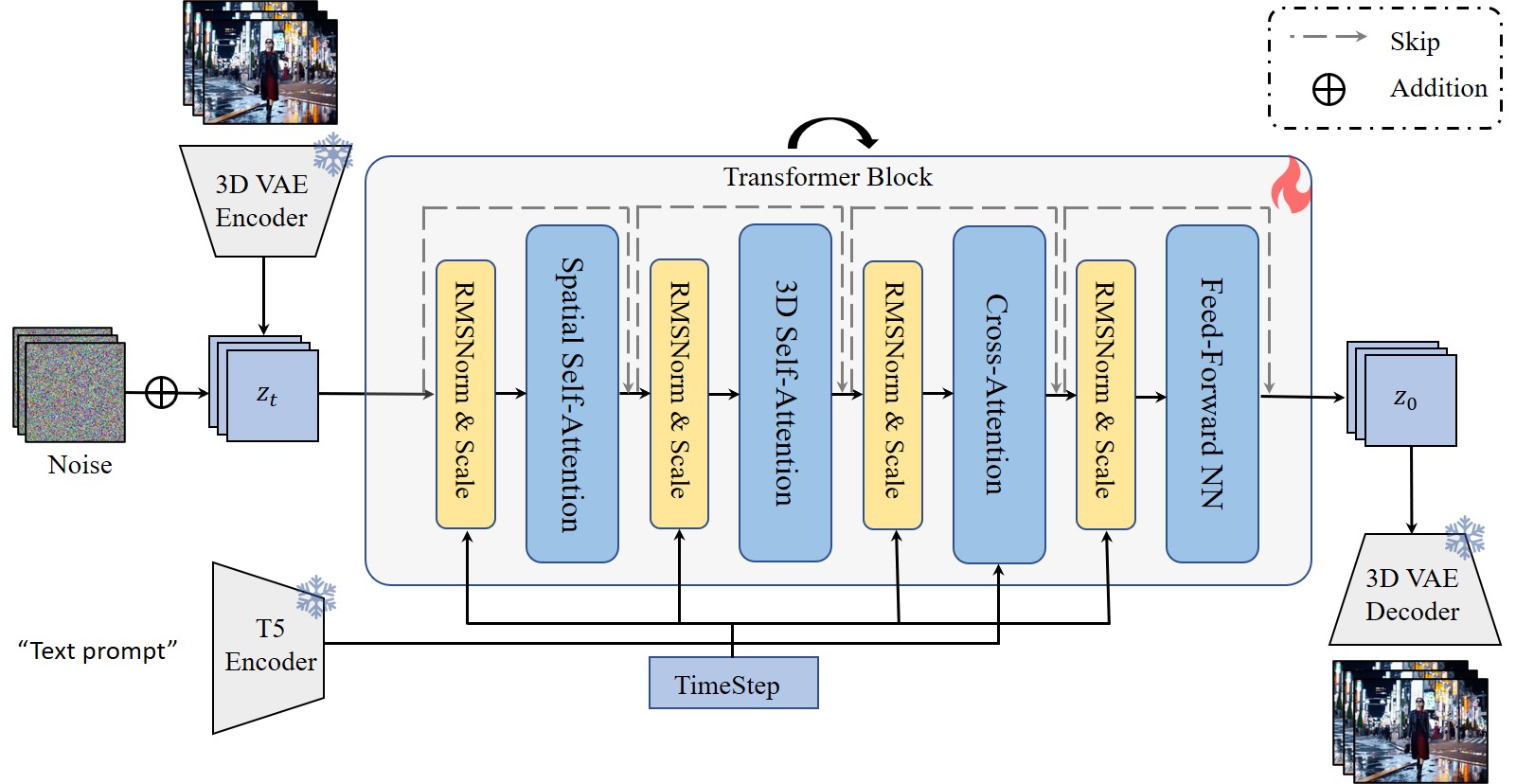}
\caption{\text{Overview of the base text-to-video generation model.}}
    \label{fig_basemodel}
\end{figure*}
\section{Introduction of the Base Text-to-Video Generation Model}
We use a transformer-based latent diffusion model as the base T2V generation model, as illustrated in Fig. \ref{fig_basemodel}. We employ a 3D-VAE to transform videos from the pixel space to a latent space, upon which we construct a transformer-based video diffusion model. Unlike previous models that rely on UNets or transformers, which typically incorporate an additional 1D temporal attention module for video generation, such spatially-temporally separated designs do not yield optimal results. We replace the 1D temporal attention with 3D self-attention, enabling the model to effectively perceive and process spatiotemporal tokens, thereby achieving a high-quality and coherent video generation model. Specifically, before each attention or feed-forward network (FFN) module, we map the timestep to a scale, thereby applying RMSNorm to the spatiotemporal tokens.
\section{Details of Collected Dataset}

In this section, we provide a detailed description of the rendered dataset used to train our model.

\paragraph{3D Environments} We collect 12 different 3D environments assets from \url{https://www.fab.com/}. To minimize the domain gap between rendered data and real-world videos, we primarily select visually realistic 3D scenes, while choosing a few stylized or surreal 3D scenes as a supplement. To ensure data diversity, the selected scenes cover a variety of indoor and outdoor settings, such as city streets, shopping malls, and the countryside.

\paragraph{Camera Trajectories} To create data that roam within a scene, we employ smoothed polylines as camera trajectories. Specifically, we begin by randomly sampling coordinate points in the 3D scene to serve as the endpoints of the polyline, and then generate B-spline curves from these points. To ensure smooth camera movement without abrupt speed changes or rotations, we limit the camera's movement distance to the range of [3m, 6m] for each 77-frame video segment and restrict the rotation angle within the xy-plane to less than 60 degrees.

Upon completing the 3D scene collection and trajectory design, we utilized Unreal Engine 5 to batch-render 100 long videos for training. Each video features 7,601 frames (30 fps) of continuous camera movement. Additionally, we record the camera's extrinsic and intrinsic parameters for each frame. The camera is configured with a focal length of 24mm, an aperture of 10, and a field of view (FOV) of 52.67 degrees.

\section{Additional Open-Domain Results}

\begin{figure*}[ht]
  \centering
  \includegraphics[width=1\linewidth]{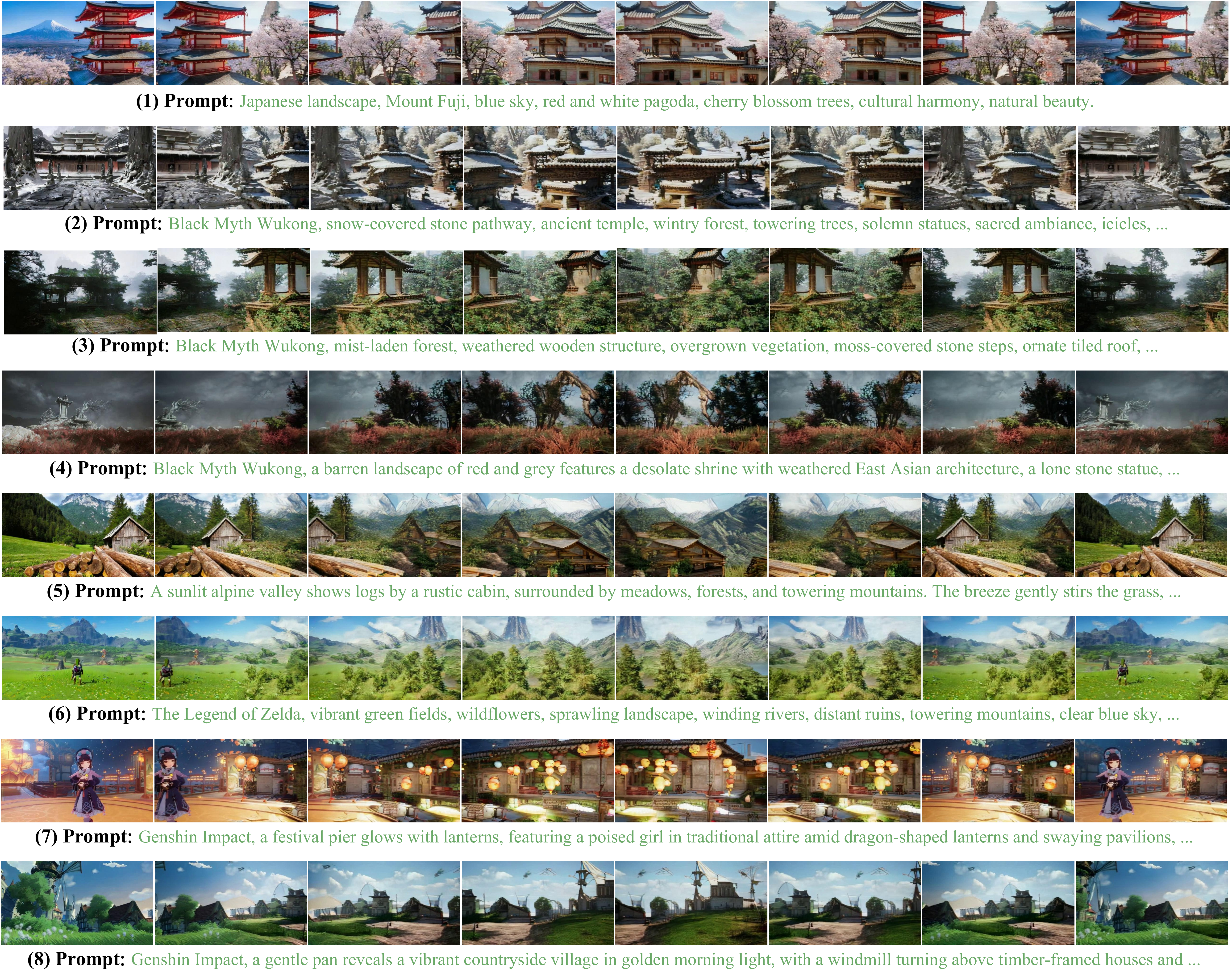}
  \caption{\textbf{Open-Domain Results.}}
\label{fig:open_domain_1} 
\end{figure*}

\begin{figure*}[ht]
  \centering
  \includegraphics[width=1\linewidth]{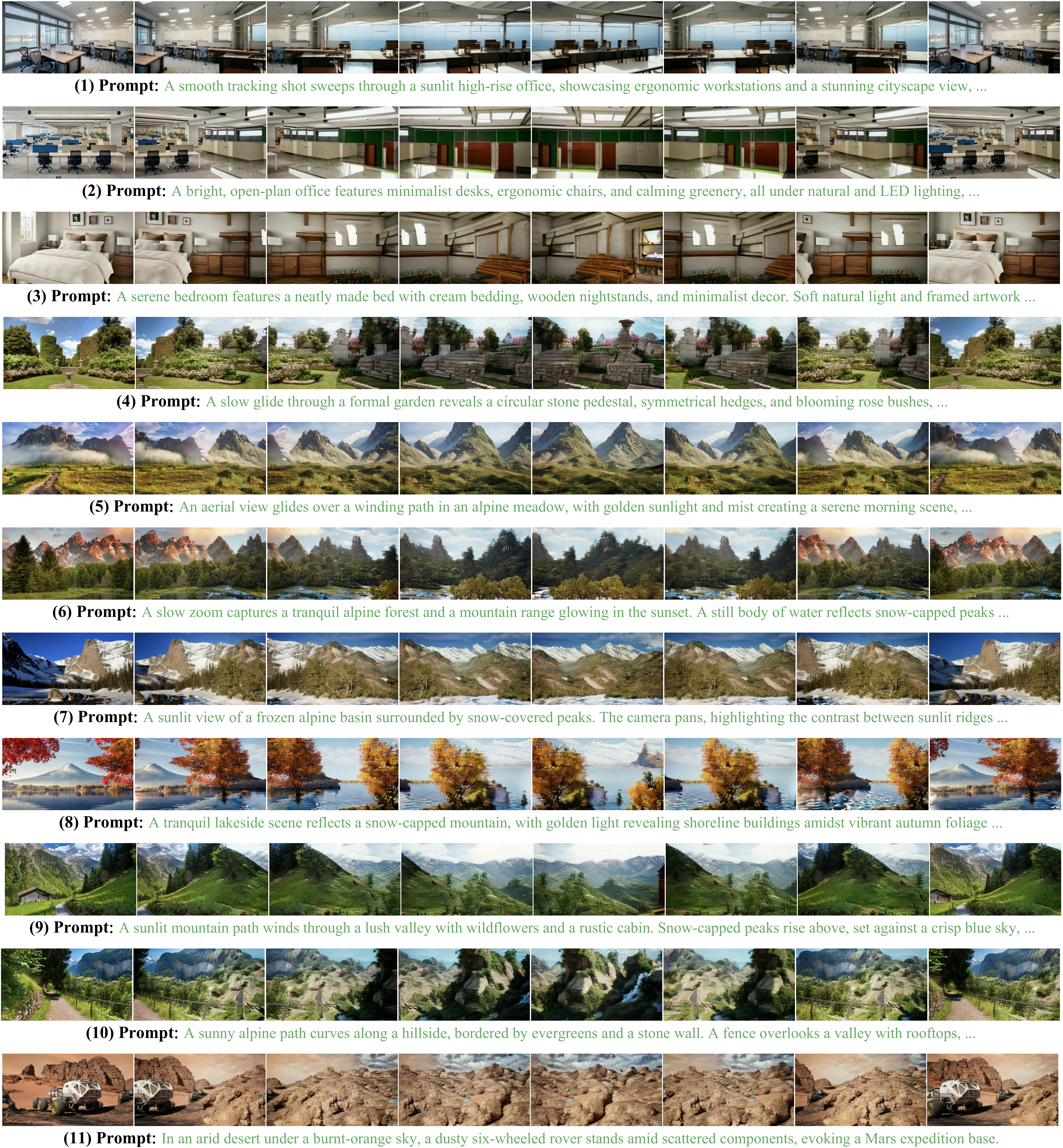}
  \caption{\textbf{Open-Domain Results.}}
\label{fig:open_domain_2} 
\end{figure*}

In Fig.~\ref{fig:open_domain_1} and Fig.~\ref{fig:open_domain_2}, we present additional open-domain results. Using diverse images collected from the internet as initial frames, we demonstrate long video generation with "rotate away and rotate back" trajectories. These source images, representing various styles and scenes, can be found in the provided Data.

Our method achieves generalization capability in open-domain scenarios due to two main factors:
(1) Training on diverse scenes enables the model to develop generalizable context utilization skills;
(2) The pre-trained base model possesses strong generative priors from exposure to various data types during pre-training.

However, our method still faces significant limitations in open-domain generalization that require future research:
(1) The 1B-parameter base model's capabilities are insufficient, only showing good results on simple trajectories. For complex trajectories, the base model struggles to generate high-quality content from the initial frame, leading to unacceptable error accumulation in long video generation. Validating our approach with larger-scale base models remains a future research direction.
(2) The method cannot yet support more complex, diverse, and dynamic long-term scene exploration in open-domain settings. Our ideal goal is to enable free, extended navigation from any given image while maintaining memory consistency. This is a challenging objective, though the "context as memory" concept shows promise.


\end{document}